\documentclass[runningheads]{llncs}
\usepackage{graphicx}
\usepackage{caption}
\usepackage{subfigure}
\usepackage{tikz}
\usepackage{comment}
\usepackage{amsmath,amssymb} % define this before the line numbering.
\usepackage{color}
\usepackage{multirow}
\usepackage{bm}
\usepackage{booktabs}
\usepackage{array}
\usepackage{wrapfig}
\usepackage[colorlinks,linkcolor=red]{hyperref}

\usepackage[capitalize]{cleveref}
\crefname{section}{Sec.}{Secs.}
\Crefname{section}{Section}{Sections}
\Crefname{table}{Table}{Tables}
\crefname{table}{Tab.}{Tabs.}

\usepackage[accsupp]{axessibility}  % Improves PDF readability for those with disabilities.

\begin{document}
% \linenumbers
\pagestyle{headings}
\mainmatter
\def\ECCVSubNumber{4919}  % Insert your submission number here

\title{KXNet: A Model-Driven Deep Neural Network for Blind Super-Resolution} % Replace with your title

\begin{comment}
\titlerunning{ECCV-22 submission ID \ECCVSubNumber} 
\authorrunning{ECCV-22 submission ID \ECCVSubNumber} 
\author{Anonymous ECCV submission}
\institute{Paper ID \ECCVSubNumber}
\end{comment}
%******************

% \begin{comment}
\author{Jiahong Fu\inst{1} \and
Hong Wang\inst{2} \and Qi Xie\inst{*1} \and Qian Zhao\inst{1} \and Deyu Meng\inst{1, 3} \and \\ Zongben Xu\inst{1, 3}}
\authorrunning{Jiahong Fu et al.}
\institute{Xi'an Jiaotong University, Shaanxi, P.R. China\\
\email{jiahongfu@stu.xjtu.edu.cn, \{xie.qi, timmy.zhaoqian, dymeng, zbxu\}@mail.xjtu.edu.cn} \and
Tencent Jarvis Lab, Shenzhen, P.R. China\\
\email{hazelhwang@tencent.com} \and 
Pazhou Lab, Guangzhou, P.R. China}
% \end{comment}
%******************
\maketitle
\renewcommand{\thefootnote}{\fnsymbol{footnote}}
\footnotetext[0]{\hspace{-2mm}* Corresponding author.}
\renewcommand{\thefootnote}{\arabic{footnote}}
\thispagestyle{empty}

\begin{abstract}
Although current deep learning-based methods have gained promising performance in the blind single image super-resolution (SISR) task,  most of them mainly focus on heuristically constructing diverse network architectures and put less emphasis on the explicit embedding of the physical generation mechanism between blur kernels and high-resolution (HR) images. To alleviate this issue, we propose a model-driven deep neural network, called KXNet, for blind SISR. Specifically, to solve the classical SISR model, we propose a simple-yet-effective iterative algorithm. Then by unfolding the involved iterative steps into the corresponding network module, we naturally construct the KXNet. The main specificity of the proposed KXNet is that the entire learning process is fully and explicitly integrated with the inherent physical mechanism underlying this SISR task. Thus, the learned blur kernel has clear physical patterns and the mutually iterative process between blur kernel and HR image can soundly guide the KXNet to be evolved in the right direction. Extensive experiments on synthetic and real data finely demonstrate the superior accuracy and generality of our method beyond the current representative state-of-the-art blind SISR methods. Code is available at: \url{https://github.com/jiahong-fu/KXNet}.
\end{abstract}

\section{Introduction}
Single image super-resolution (SISR) has been widely adopted in various vision applications,  \emph{e.g.} video surveillance, medical imaging, and video enhancement. For this SISR task, the main goal is to reconstruct the high-resolution (HR) image with high visual quality from an observed low-resolution (LR) image.

Specifically, in traditional SISR framework, the degradation process for an LR image $\bm{Y}$ can be mathematically expressed as \cite{elad1997restoration,farsiu2004advances}:
\begin{equation}
  {\bm{Y} = (\bm{X} \otimes \bm{K}) \downarrow _{{\mathbf{s}}} + \bm{N}},
  \label{eq:important}
\end{equation}
\noindent
where $\bm{X}$ is the to-be-estimated HR image; $\bm{K}$ is a blur kernel; $\otimes$ denotes two-dimensional (2D) convolution operation;  $\downarrow _{{\mathbf{s}}}$ represents the standard $s$-fold downsampler, \textit{i.e.}, only keeping the upper-left pixel for each distinct ${s\times s}$ patch \cite{zhang2020deep}; $\bm{N}$ denotes the Additive White Gaussian Noise (AWGN) with noise level $\sigma$. Clearly, estimating $\bm{X}$ and $\bm{K}$ from $\bm{Y}$ is an ill-posed inversion problem.

With the rapid development of deep neural networks (DNNs), in recent years,  many deep learning (DL)-methods have been proposed for this SISR
task~\cite{kim2016deeply,lai2017deep,lim2017enhanced,zhang2018residual,zhang2018image,niu2020single,liang2021swinir}. Albeit achieving promising performance in some scenes, the assumption that the blur kernel $\bm{K}$ is known, such as bicubic \cite{dong2015image,kim2016deeply,lim2017enhanced,zhang2018image,zhang2018residual}, would make these methods tend to fail in real applications where the practical degradation process is always complicated.
% In specific, various researchers have assumed that the blur kernel in \cref{eq:important} is known, such as bicubic \cite{dong2015image,kim2016deeply,lim2017enhanced,zhang2018image,zhang2018residual}. Albeit achieving promising performance in some scenes, these methods often tend to fail in real applications where the pre-specified assumption does not well comply with the practical degradation. 
To alleviate this issue, researchers have focused on the more challenging blind super-resolution (SR) task where the blur kernel $\bm{K}$ is unknown. Currently, blind SR methods can be mainly divided into two categories: traditional-model-based ones and DL-based ones. 

Specifically, conventional blind SR works \cite{marquina2008image,dai2009softcuts} aim to formulate the hand-crafted prior knowledge of blur kernel $\bm{K}$ and HR image $\bm{X}$, into an optimization algorithm to constrain the solution space of the ill-posed SISR problem. Due to the involved iterative computations, these methods are generally time-consuming. Besides, the manually-designed priors cannot always sufficiently represent the complicated and diverse images in real scenarios.

Recently, to flexibly deal with multi-degradation situations, some DL-based blind SR methods \cite{zhang2019deep,bell2019blind,liang2021flow} have been proposed, which are composed of two successive steps, \textit{i.e.,} blur kernel estimation and non-blind super-resolver. Since these two steps are independently handled, the estimated blur kernel and the recovered HR image are possibly not compatible well. To further boost the performance,
some works~\cite{gu2019blind,luo2020unfolding,wang2021unsupervised,zhang2021designing,wang2021real} directly utilized off-the-shelf network modules to recover HR images in an end-to-end manner without fully embedding the physical generation mechanism underlying this SISR task.

Very recently, the end-to-end deep unfolding framework has achieved good performance in this SISR task~\cite{heide2016proximal,brifman2019unified,zhang2020deep,luo2020unfolding}. Typically,  by alternately updating the blur kernel $\bm{K}$ and the HR image $\bm{X}$, the blind SR work~\cite{luo2020unfolding} heuristically constructs an optimization-inspired SR network. However, there are two main limitations: 1) the inherent physical generation mechanism in \cref{eq:important} is still not fully and explicitly embedded into the iterative computations of $\bm{K}$ and $\bm{X}$, and every network module has relatively weak physical meanings; 2) most of these deep unfolding-based methods cannot finely extract blur kernels with clear physical patterns, which is caused by weak interpretable operations on blur kernels, such as concatenation and stretching \cite{luo2020unfolding} and spatial feature transform (SFT) layer~\cite{gu2019blind}. Hence, there is still room for further performance improvement.

To alleviate these issues, we propose a novel deep unfolding blind SR network that explicitly embeds the physical mechanism in \cref{eq:important} into the mutual learning between blur kernel and HR image in a sufficient manner. Under the explicit guidance of the degradation process \cref{eq:important}, the updating of $\bm{K}$ and $\bm{X}$ finely proceeds, and the learned blur kernel presents clear structural patterns with specific physical meanings. In summary, our contributions are mainly three-fold:

%-------------------------------
\begin{figure}[t]
    \centering
    \setlength{\abovecaptionskip}{0.1cm}
    \includegraphics[width=0.8\linewidth]{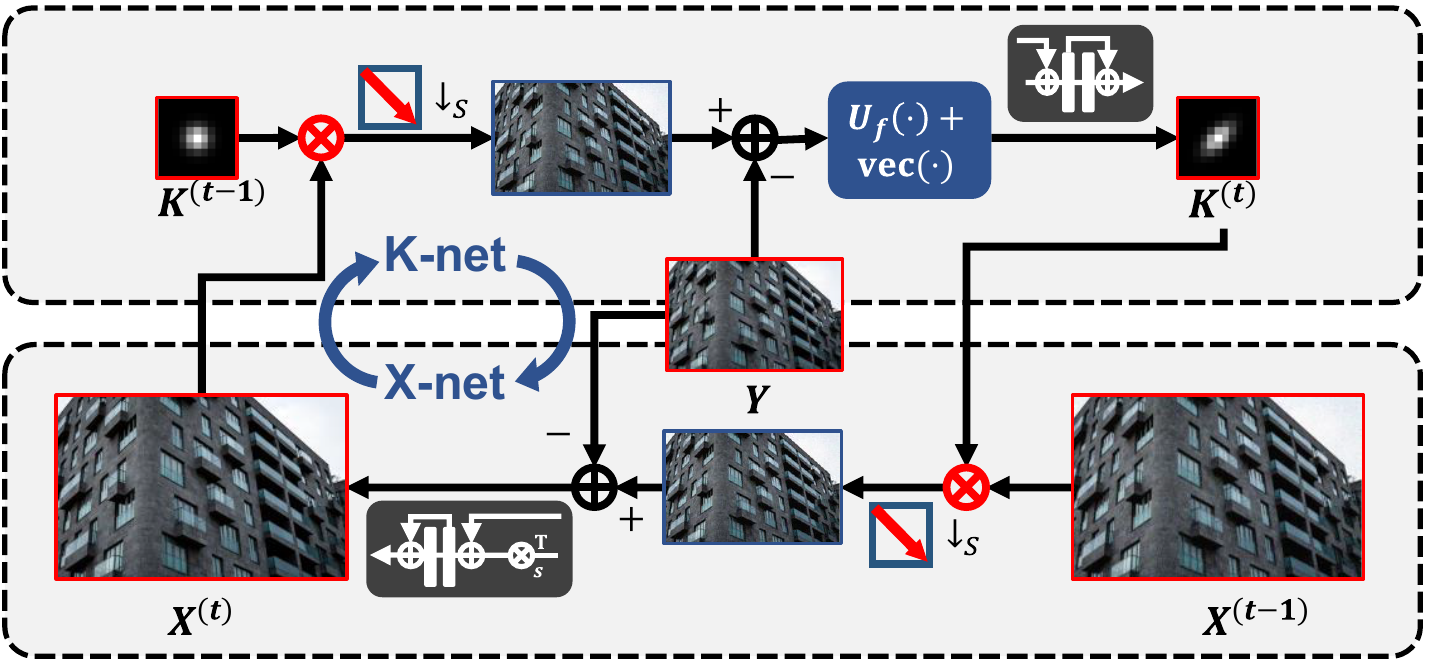}
    \caption{Illustration of the proposed KXnet where K-net for blur kernel estimation and X-net for HR image estimation are constructed based on the physical generation mechanism in \cref{eq:important}.}
    \label{fig:figure1}
    \vspace{-0.4cm}
\end{figure}
%---------------------------------

\begin{itemize}
\item We propose a novel model-driven deep unfolding blind super-resolution network (called KXNet) to jointly estimate the blur kernel $\bm{K}$ and the HR image $\bm{X}$, which is explicitly integrated with the physical generation mechanism in \cref{eq:important}, as shown in \cref{fig:figure1}. Specifically, we propose an iterative algorithm to solve the classical degradation model \cref{eq:important} and then construct the KXNet by unfolding the iterative steps into the corresponding network modules. Naturally, the mutually iterative learning process between blur kernel and HR image fully complies with the inherent physical generation mechanism, and every network module in KXNet has clear physical interpretability. 

\item Instead of heuristic operations (\emph{e.g.} concatenation and affine transformation) on blur kernel in most of the current SR methods, the learning and estimation of blur kernel in our method is proceeding under the guidance of \cref{eq:important} and thus has clearer physical meanings. As shown in \cref{fig:figure1}, the K-net is finely corresponding to the iterative steps for updating blur kernel and thus the extracted blur kernel $\bm{K}$ has reasonable and clear physical structures. Besides, attributed to the intrinsic embedding of the physical generation mechanism, we maintain the essential convolution computation between blur kernel and HR image, which is expected to achieve better SR performance.

% Instead of heuristically estimating and utilizing the blur kernel as in the current most SR methods, the estimation and the usage of the blur kernel in the proposed method have clearer physical meanings and 
% Specifically, as shown in \cref{fig:figure1}, the K-net is corresponding to the iterative process of the proposed unfolding algorithm and the extracted blur kernel has reasonable and clear physical structures. 

% Besides, during the iterative process, we maintain the essential convolution operation in \cref{eq:important} to better estimate the HR image, which is finely following the traditional SISR framework.

\item Extensive experiments executed on synthetic and real data comprehensively demonstrate the superiority of the proposed KXNet in SR performance and model generalizability beyond the current state-of-the-art (SOTA) methods. Besides, more analysis and network visualization validate the rationality and effectiveness of our method, and the extracted blur kernels with clear structures would be helpful for other vision tasks in real applications.
% More analysis and network visualization validate the effectiveness and the generality of the proposed KXNet. Besides, the estimated blur kernel with clear patterns should be useful for other related tasks.
\end{itemize}
%------------------------------------------------------------------------

\section{Related Work}
\subsection{Non-Blind Single Image Super-Resolution}

In recent years, deep learning (DL) has achieved great progress in SISR task. Current SISR methods mainly focus on utilizing deep neural networks to learn the mapping function from a low-resolution (LR) image to the corresponding high-resolution (HR) image via paired training data.  Since it is very time-consuming and labor-intensive to pre-collect massive paired LR-HR images, many researchers adopt the manually-designed degradation processes to generate the LR images, such as the classical bicubic interpolation ($\bm{K}$ in \cref{eq:important} is set as the bicubic kernel). This setting has been widely adopted from the early SRCNN \cite{dong2015image} to the recent various SISR methods \cite{lim2017enhanced,zhang2018image,niu2020single,liang2021swinir}.  These methods aim to design diverse network modules to improve the SISR performance.

Considering that in real scenes, the degradation process is always complicated, there are some works \cite{zhang2018learning,xu2020unified,zhang2020deep} dealing with multiple degradation forms. For example, SRMD \cite{zhang2018learning} takes different degradation feature maps as additional inputs for the SR task. Very recently, Zhang \cite{zhang2020deep} constructs an optimization-inspired non-blind SISR network for handling the multiple degradation scenes.

\subsection{Blind Single Image Super-Resolution}

To better represent the real degradation process and improve the SR performance in real-world, blind single image super-resolution has been attracting the attention of researchers in this field. In this case, the goal is to jointly estimate the unknown blur kernel $\bm{K}$ and the expected HR image $\bm{X}$. Currently, against this task, the existing methods can be mainly categorized into two groups: two-step methods and end-to-end methods.

\noindent
{\bf Two-step Blind Super-Resolution.} In this research line, researchers first estimate the blur kernel based on different prior knowledge \cite{pan2016blind,yan2017image,ren2020neural,liang2021flow}. For example, Michaeli \cite{michaeli2013nonparametric} utilizes the inter-scale recurrence property of an image to help extract the blur kernel. Then by inserting the estimated blur kernel into non-blind SR methods\cite{zhang2019deep}, the corresponding SR results can be restored. Recently, Kligler \cite{bell2019blind} have proposed an unsupervised KernelGAN to estimate the blur kernel based on the recurrence property of the image patch, then utilized the extracted kernel to help reconstruct SR images. Similarly, Liang \cite{liang2021flow} have proposed a flow-based architecture to capture the prior knowledge of the blur kernel which can be used for the subsequent non-blind SR task. Most of these two-step methods have not fully considered the iterative and mutual learning between blur kernels and HR images.

\noindent
{\bf End-to-End Blind Super-Resolution.} 
Very recently, some works begin to emphasize how to merge the kernel estimation process with the non-blind SR process and thus design an end-to-end blind SR framework. 
Gu \cite{gu2019blind} firstly designed a single network architecture that contains a kernel estimation module and a non-blind SR module. However, this method needs to separately train multiple modules. To alleviate this issue,  Luo~\cite{luo2020unfolding,luo2021end} proposed a complete end-to-end network that can be trained in an end-to-end manner. Wang \cite{wang2021unsupervised} proposed an unsupervised degradation representation learning scheme and then utilized it to help accomplish the blind SR task. Zhang \cite{zhang2021designing} and Wang \cite{wang2021real} design a ``high-order" degradation model to simulate the image degradation process.

Albeit achieving promising performance in some scenarios, most of these methods have the following limitations: 1) The estimated degradation form has relatively weak physical meanings or cannot fully reflect blur kernels with clear and structural patterns; 2) The degradation representation is heuristically used, such as simple concatenation with LR image,  without explicitly reflecting the inherent convolution operation between blur kernel and image; 3) The intrinsic physical generation mechanism in \cref{eq:important} has not been fully embedded into network design.To alleviate these issues, we adopt the deep unfolding technique \cite{xie2019multispectral,wang2020model}, with several novel designs, for better embedding the inherent mechanism into network structure and improving the performance in blind SR restoration.

%--------------------------------------------------

%-------------------------------------------------------------------------

%-------------------------------------------------------------------------
\section{Blind Single Image Super-Resolution Model}

In this section, for the blind SISR task, we formulate the corresponding mathematical model and propose an optimization algorithm.

% In this section, we first derive the mathematical model for the blind SISR task and then propose the corresponding optimization algorithm.

\subsection{Model Formulation}
From \cref{eq:important}, given an observed LR image $\bm{Y}\in \mathbb{R}^{h \times w}$, our goal is to estimate the two unknown variables, \textit{i.e.}, blur kernel $\bm{K}\in \mathbb{R}^{p \times p}$ and HR image $\bm{X}\in \mathbb{R}^{H \times W}$. Correspondingly, we can formulate the following optimization problem as:
\begin{equation}
\small
  \begin{split}
      &\operatorname*{min}_{\bm{K}, \bm{X}} \Big\| \bm{Y} - \left( \bm{X} \otimes \bm{K} \right) \downarrow_{\mathbf{s}}  \Big\| ^2_F + \lambda_1 \phi_1 (\bm{K}) + \lambda_2 \phi_2 (\bm{X}) \\
      & s.t. ~~ \bm{K}_j \geq 0, \sum_{j} \bm{K}_j = 1, \forall j,
  \end{split}
  \label{eq:important2}
\end{equation}
% \normalsize
where $\phi_1 (\bm{K})$ and $\phi_2 (\bm{X})$ represent the regularizers for delivering the prior knowledge of blur kernel and HR image, respectively; $\lambda_1$ and $\lambda_2$ are trade-off regularization parameters.  Similar to \cite{perrone2014total,sun2013edge,ren2020neural}, we introduce the non-negative and equality constraints for every element $\bm{K}_j$ in blur kernel $\bm{K}$. Specifically, the data fidelity term (\emph{i.e.}, the first term in the objective function of~\cref{eq:important2}) represents the physical generation mechanism, which would provide the explicit guidance during the iterative updating of $\bm{K}$ and $\bm{X}$, and the prior terms $\phi_1 (\bm{K})$ and $\phi_2 (\bm{X})$ enforce the expected structures of the solution for this ill-posed problem.

Instead of adopting hand-crafted prior functions as in conventional optimization-based SR methods, we utilize a data-driven strategy to flexibly extract the implicit prior knowledge underlying $\bm X$ and $\bm K$ from data via DNNs in an end-to-end manner. This operation has been fully validated to be effective in many diverse vision tasks by extensive studies~\cite{zhang2017learning,xie2019multispectral,wang2020model}. The details for learning $\phi_1 (\bm{K})$ and $\phi_2 (\bm{X})$ are given in Sec.~\ref{sec:net}.

\subsection{Model Optimization}\label{sec:opt}
% To solve the SR problem in \cref{eq:important2}, current researchers often adopt the alternative iterative optimization strategy~\cite{luo2020unfolding, luo2021end} to heuristically construct the deep network frameworks based on the off-the-shelf network modules.

\begin{wrapfigure}{r}{6cm}
\vspace{-9mm}
\centering
\includegraphics[width=6.1cm]{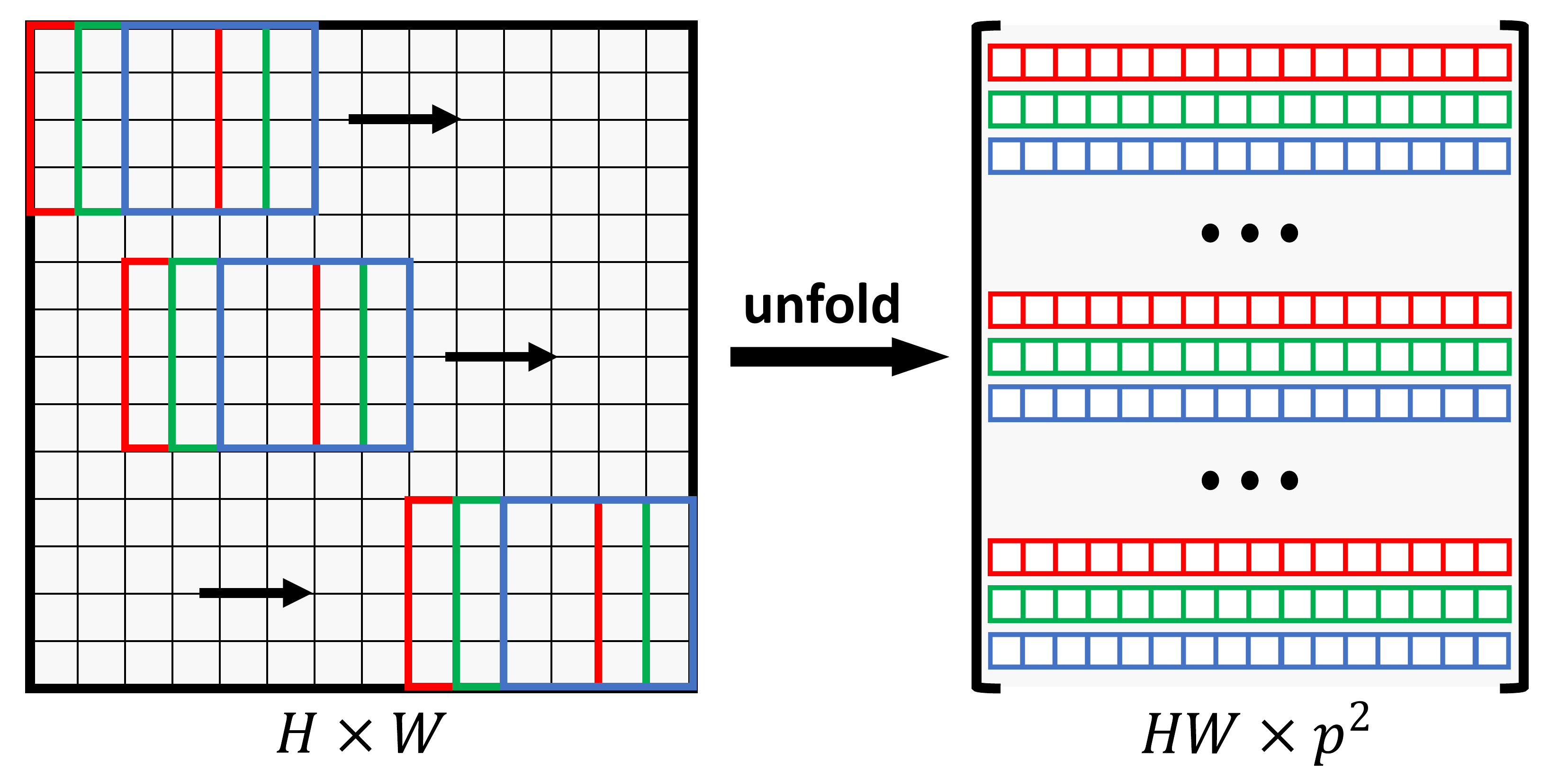}
\vspace{-2mm}
  \caption{Illustration of the operator $U_f (\cdot)$.}
\label{fig:unfold}
\vspace{-1mm}
\end{wrapfigure}

For this blind SISR task, our goal is to build a deep unfolding network where network modules are possibly corresponding to iterative steps involved in an optimization algorithm so as to make the network interpretable and easily controllable. Thus, it is necessary to derive an iterative algorithm for solving the SR problem in \cref{eq:important2}. To this end, we adopt a proximal gradient technique \cite{beck2009fast,gregor2010learning} to alternatively update the blur kernel $\bm{K}$ and HR image $\bm{X}$. Then, the derived optimization algorithm contains only simple operators which makes it possible to be easily unfolded into network modules, as shown in following.
 
\vspace{1mm}
\noindent
{\bf Updating blur kernel $\bm{K}$}: The blur kernel $\bm{K}$ can be optimized by solving the quadratic approximation \cite{beck2009fast} of the problem \cref{eq:important2} with respect to the variable $\bm{K}$, expressed as:
\begin{equation}
\small
  \begin{split}
      &\operatorname*{min}_{\bm{K}} \Big\| \bm{K} \!- \!\left(\! \bm{K}^{(t-1)}
      \!-\! \delta_{1} \nabla f \left(\bm{K}^{(t-1)}\!\right) \right)  \Big\| ^2_F \!+\! \ \lambda_1 \delta_1 \phi_1 (\bm{K}) \\
      & s.t. ~~ \ \bm{K}_j \geq 0, \sum_{j} \bm{K}_j = 1, \forall j,
  \end{split}
  \label{eq:important3}
  \vspace{-0.4cm}
\end{equation}
\normalsize
where $\bm{K}^{(t-1)}$ denotes the updating result after the last iteration; $\delta_1$ denotes the stepsize parameter; $f(\bm{K}^{(t-1)}) = \left \|  \bm{Y} - \left( \bm{X}^{(t-1)} \otimes \bm{K}^{(t-1)} \right) \downarrow_{\mathbf{s}} \right \|_{F}^{2}$.
For a general regularizer $\phi_1 (\cdot)$, the solution of \cref{eq:important3} can be easily expressed as \cite{donoho1995noising}:
\begin{equation}
   {\bm{K}}^{(t)} = \operatorname{prox}_{\lambda_1 \delta_1} \left( {\bm{K}}^{(t-1)} - \delta_1 \nabla f \left( {\bm{K}}^{(t-1)} \right) \right),
    \label{eq:important5}
\end{equation}
where the specific form of $\nabla f \left( {\bm{K}}^{(t-1)}\right)$ is complicated. For ease of calculation, by transforming the convolutional operation in $f(\bm{K}^{(t-1)})$ into matrix multiplication, we can derive that:

%-----------------------------------------------------------

\begin{equation}
\begin{split}
  f\!\left({\bm{k}}^{(t-1)} \right) & \!=\!\text{vec}\left(\left \|{ \bm{Y} \!- \!\left( \bm{X}^{(t-1)}\! \otimes\! \bm{K}^{(t-1)} \right) \downarrow_{\mathbf{s}}}\right \|_{F}^{2}\right)   \\
 & \!=\left \|  \bm{y} - D_{\mathbf{s}} U_f\left( \bm{X}^{(t-1)}\right) {\bm{k}}^{(t-1)}\right \|_{F}^{2},
  \end{split}
\end{equation}
where ${\bm{y}}=\text{vec}\left(\bm{Y}\right)$  and ${\bm{k}}=\text{vec}\left(\bm{K}\right)$ denote the vectorizations of $\bm{Y}$ and $\bm{K}$, respectively; $\bm{y}\in\mathbb{R}^{hw\times1}$;$\bm{k}\in\mathbb{R}^{p^2\times1}$; $U_f\left({ \bm{X}^{(t-1)}}\right)\in\mathbb{R}^{HW\times p^2}$ are the unfolded result of $\bm{X}^{(t-1)}$ (see \cref{fig:unfold}); $D_{\mathbf{s}}$ denotes the downsampling operator which is corresponding to the operator $\downarrow _{{\mathbf{s}}}$, and achieves the transformation from the size $HW$ to the size $hw$. Thus, the result $D_{\mathbf{s}} U_f\left( \bm{X}^{(t-1)}\right)$\footnote{More derivations are provided in the supplementary material.} has the size with $hw\times p^2$  and $\nabla f ({\bm{k}}^{\left(t-1\right)})$ is derived as:
% \begin{equation}
% \small
%     \nabla f({\bm{k}}^{(t-1)}) = \left( D_{\mathbf{s}} U_f\left(\bm{X}^{(t-1)}\right) \right)^\mathrm{T} \left( {\bm{y}} - D_{\mathbf{s}} U_f\left(\bm{X}^{(t-1)}\right) {\bm{k}}^{(t-1)}\right),
%     \label{eq:important6}
% \end{equation}
\begin{equation}
\small
    \nabla \!f\!\left({\bm{k}}^{(t-1)}\right)\! =\!\left(\! D_{\mathbf{s}}U_f\!\!\left(\!\bm{X}^{(t-1)}\!\right)\! \right)^\mathrm{T}\!\!\!\text{vec}\!\left(\!{ \bm{Y}\!\! - \!\!\left(\! \bm{X}^{(t-1)} \!\otimes\! \bm{K}^{(t-1)} \!\right)\! \downarrow_{\mathbf{s}}}\!\right),
    \label{eq:important6}
\end{equation}
\normalsize
where $\nabla f ({\bm{k}}^{\left(t-1\right)}) \in \mathbb{R}^{p^{2}\times 1}$; $\nabla f ({\bm{K}}^{\left(t-1\right)}) = \text{vec}^{-1}\left(\nabla f ({\bm{k}}^{\left(t-1\right)})\right)$; $\text{vec}^{-1}(\cdot)$ is the reverse vectorization; $\operatorname{prox}_{\lambda_1 \delta_1}(\cdot)$ is the proximal operator dependent on the regularization term $\phi_1 (\cdot)$ with respect to $\bm K$. Different from the traditional methods with hand-crafted regularization terms, we rely on the powerful fitting capability of residual networks to automatically learn the implicit proximal operator $\operatorname{prox}_{\lambda_1 \delta_1}(\cdot)$ via training data. Such operations have achieved great success in other deep unfolding works~\cite{xie2019multispectral,wang2020model}. The details are described in Sec.~\ref{sec:net}.

\vspace{1mm}
\noindent
{\bf Updating HR image $\bm{X}$}: Similarly, the quadratic approximation of the problem in \cref{eq:important2} with respect to $\bm{X}$ can be derived as:
\begin{equation}
\small
    \operatorname*{min}_{\bm{X}} \Big\| \bm{X} \!- \!\left( \bm{X}^{(t-1)}
      \!-\! \delta_{2} \nabla h \left(\bm{X}^{(t-1)}\right) \right)  \Big\| ^2_F \!+\! \lambda_2 \delta_2 \phi_2(\bm{X}),
    \label{eq:important7}
\end{equation}
% \normalsize
where $h\left(\bm{X}^{(t-1)}\right) = \left \|  \bm{Y} - \left( \bm{X}^{(t-1)} \otimes \bm{K}^{(t)} \right) \downarrow_{\mathbf{s}} \right \|_{F}^{2}$; With $\nabla h \left( \bm{X}^{(t-1)}\right) =  \\ \bm{K}^{(t)} \otimes^{\mathrm{T}}_{\mathbf{s}} \left( \bm{Y} - \left( \bm{X}^{(t-1)} \otimes \bm{K}^{(t)}\right) \downarrow_{\mathbf{s}}\right)$, we can deduce the updating rule for $\bm{X}$ as:
\vspace{-1mm}
\begin{equation}
\small
\hspace{-0.3mm}
    \bm{X}^{(t)} = \operatorname{prox}_{\lambda_2 \delta_2}\! \!\Big( \bm{X}^{(t-1)} \!-\! \delta_2 \bm{K}^{(t)} \!\otimes_{\mathbf{s}}^{\mathrm{T}}\! \left( \bm{Y} \!-\! (\bm{X}^{(t-1)} \!\otimes\! \bm{K}^{(t)}) \downarrow_{\mathbf{s}} \right)\!\Big),
    \label{eq:important8}
\end{equation}
% \normalsize
where $\operatorname{prox}_{\lambda_2 \delta_2}(\cdot)$ is the proximal operator dependent on the regularization term $\phi_2 (\cdot)$ about $\bm{X}$; $\otimes_{\mathbf{s}}^{\mathrm{T}}$ denotes the transposed convolution operation with stride as $\mathbf{s}$. Similar to  $\operatorname{prox}_{\lambda_1 \delta_1}(\cdot)$, we adopt deep network to flexibly learn the $\operatorname{prox}_{\lambda_2 \delta_2}(\cdot)$.

As seen, the proposed optimization algorithm is composed of the iterative rules \cref{eq:important5} and \cref{eq:important8}. By unfolding every iterative step into the corresponding network module, we can naturally build the deep unfolding network architecture for solving the blind SISR task as given in~\cref{eq:important2}.

%-------------------------------------------------------------------------

\begin{figure*}[t]
\centering
\subfigure[The algorithm processes of the entire KXNet.]{
    \label{fig:fig2_a}
    \begin{minipage}[b]{0.99\linewidth}
    \centerline{\includegraphics[width=12.5cm]{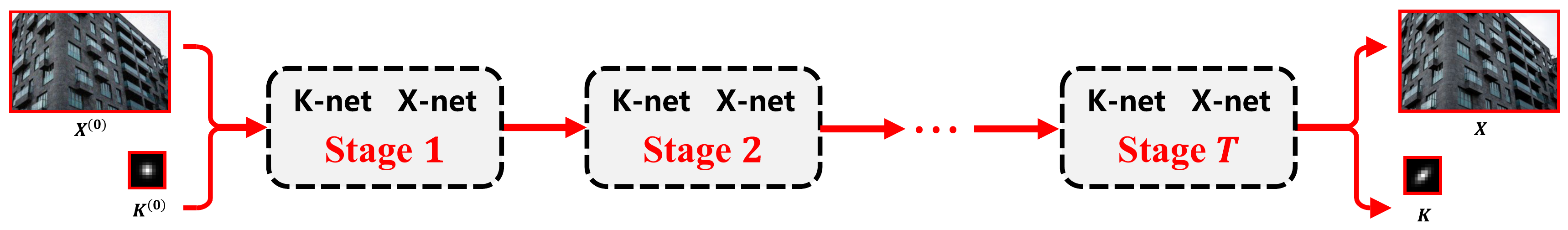}}
    \end{minipage}
    % \label{fig:fig2_a}
    } %[图片大小]{图片路径}
\subfigure[The network architecture of K-net and X-net at each stage.]{
    \label{fig:fig2_b}
    \begin{minipage}[b]{0.99\linewidth}
    \centerline{\includegraphics[width=12.5cm]{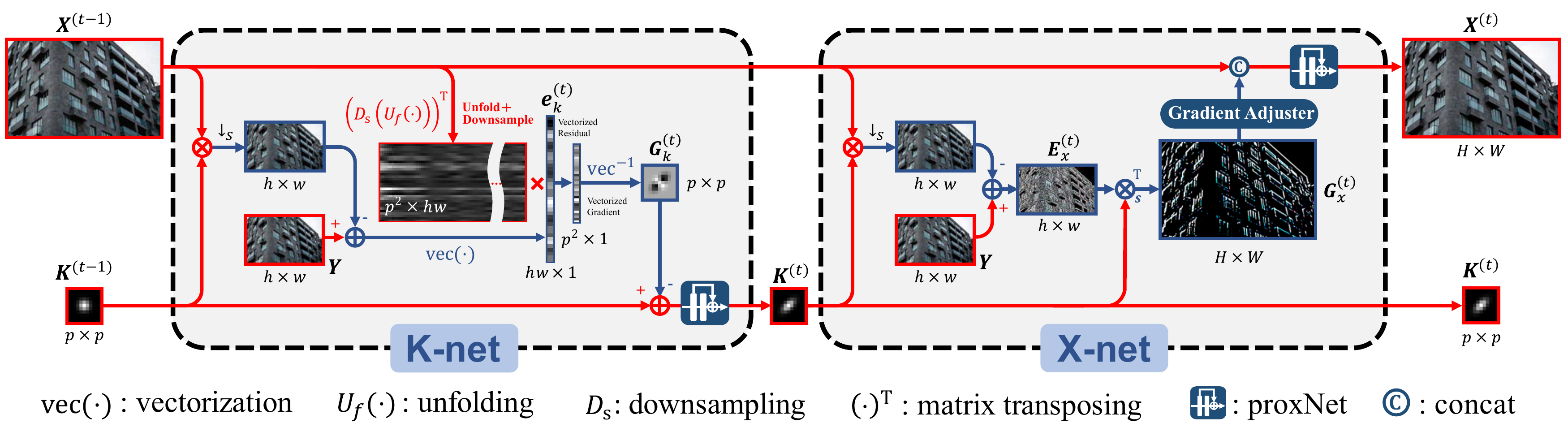}}
    \end{minipage}
    % \label{fig:fig2_b}
    } %[图片大小]{图片路径}
\vspace{-2mm}
\caption{(a) The overall architecture of the proposed KXNet contains T stages. It inputs the initialized HR image $\bm X^{(0)}$ and initialized blur kernel $\bm K^{(0)}$ , and outputs the HR image $\bm X$ and the blur kernel $\bm K$. (b) The network architecture at the $t^{\text{th}}$ stage, which consists of K-net and X-net for updating the blur kernel $\bm K$ and HR image $\bm X$, respectively.} %图片标题
\label{fig:short}  %图片交叉引用时的标签
\vspace{-4mm}
\end{figure*}

\section{Blind Super-Resolution Unfolding Network}\label{sec:net}

Recently, deep unfolding techniques have achieved great progress in various computer vision fields~\cite{yang2016deep,yang2018proximal,wang2021dicdnet,wang2021indudonet,wang2021rcdnet}, such as spectral image fusion~\cite{xie2019multispectral,xie2020mhf}, deraining~\cite{wang2020model}, and non-blind super-resolution~\cite{zhang2020deep}. Inspired by these methods, in this section, we aim to build an end-to-end deep unfolding network for blind super-resolution problem by unfolding each iterative step involved in \cref{eq:important5} and \cref{eq:important8} as the corresponding network module.

As shown in \cref{fig:fig2_a}, the proposed network consists of $T$ stages, which correspond to $T$ iterations of the proposed optimization algorithm for solving the problem in \cref{eq:important2}. At each stage, as illustrated in \cref{fig:fig2_b}, the network is subsequently composed of K-net and X-net. In specific, the K-net takes the LR image $\bm{Y}$, the estimated blur kernel $\bm{K}^{(t-1)}$ and the estimated HR image $\bm{X}^{(t-1)}$ as inputs and outputs the updated $\bm{K}^{(t)}$. Then, X-net takes $\bm{Y}$, $\bm{K}^{(t)}$, and $\bm{X}^{(t-1)}$ as inputs, and outputs the updated $\bm{X}^{(t)}$. This alternative iterative process complies with the proposed algorithm.

% \vspace{-6mm}
\subsection{Network Module Design}
By step-by-step decomposing the iterative rules of \cref{eq:important5} and \cref{eq:important8} into sub-steps and then unfolding them as the fundamental network modules, we can easily construct the entire deep unfolding framework. However, the key problem is how to deal with the implicit proximal operators $\operatorname{prox}_{\lambda_1 \delta_1}(\cdot)$ and $\operatorname{prox}_{\lambda_2 \delta_2}(\cdot)$. As stated in
Sec.~\ref{sec:opt}, following the current other unfolding-based networks~\cite{xie2019multispectral,wang2020model}, we can utilize ResNet \cite{he2016deep} to construct $\operatorname{prox}_{\lambda_1 \delta_1}(\cdot)$ and $\operatorname{prox}_{\lambda_2 \delta_2}(\cdot)$. Thus, at the $t^{\text{th}}$ stage, the network is built as:
\small
\begin{equation}
  \text{$K$-net:}\left\{
            \begin{aligned}
            % & \bm{k}^{(t-1)} =\text{vec}\left(\bm{K}^{(t-1)}\right)\\
            % & \bm{y} = \text{vec}\left(\bm{Y}\right)\\
            % &  \text{vec}\left(\left \|{ \bm{Y} - \left( \bm{X} \otimes \bm{K} \right) \downarrow_{\mathbf{s}}}\right \|_{F}^{2}\right)
            &\bm{e}_k^{(t)} =  \text{vec}\left({ \bm{Y} - \left( \bm{X}^{(t-1)} \otimes \bm{K}^{(t-1)} \right) \downarrow_{\mathbf{s}}}\right)\\
            &\bm{G}_{k}^{(t)} =\text{vec}^{-1}\!\left( \left(D_{\mathbf{s}} U_f\left(\bm{X}^{(t-1)}\right) \right)^{\mathrm{T}}\bm{e}_k^{(t)}\right) \\
            &\bm{K}^{(t)} = \operatorname{proxNet}_{\theta_{k}^{(t)}} \! \left(\bm{K}^{(t-1)} \!-\! \delta_1  \left(\bm{G}_{k}^{(t)} \right) \right),\\
            % &\bm{K}^{(t)} = \text{vec}^{-1}\left(\operatorname{proxNet}_{\theta_{k}^{(t)}} \left( \bm{k}^{(t-1)} - \bm{g}_{k}^{(t)} \right)\right),
            %& \bm{K}^{(t)} = \text{vec}^{-1}\left(\bm{k}^{(t)}\right),
            \end{aligned}
    \right.
    \label{eq:important9}
\end{equation}
\begin{equation}
    % \hspace{-7mm} 
    \text{$X$-net:}\left\{
            \begin{aligned}
            &\bm{E}_x^{(t)} = \bm{Y} - (\bm{X}^{(t-1)} \otimes \bm{K}^{(t)}) \downarrow_{\mathbf{s}} \\
            &\bm{G}_x^{(t)} = \bm{K}^{(t)} \otimes_{\mathbf{s}}^{\mathrm{T}} \bm{E}_x^{(t)} \\
            &\hat{\bm{G}}_x^{(t)} = \operatorname{adjuster}\left(\bm{G}_x^{(t)}\right) \\
            &\bm{X}^{(t)} = \operatorname{proxNet}_{\theta_x^{(t)}} \left( \bm{X}^{(t-1)}, \hat{\bm{G}}_x^{(t)} \right),
            \end{aligned}
    \right.
    \label{eq:important10}
\end{equation}
\normalsize
where $\operatorname{proxNet}_{\theta_k^{(t)}}$ and $\operatorname{proxNet}_{\theta_x^{(t)}}$ are two shallow ResNets with the parameters $\theta_k^{(t)}$ and $\theta_x^{(t)}$ at the $t$-th stage, respectively; $\operatorname{adjuster}(\cdot)$ is an operation for boosting the gradient, whose details are discussed later. All these network parameters can be automatically learned from training data in an end-to-end manner. Note that the proximal gradient descent algorithm usually needs lots of iterations for convergence, which will lead to too many network stages when adopting the unfolding technique. To avoid this issue, as shown in the last equation of \cref{eq:important10}, instead of directly adopting the subtraction between $\bm{X}^{(t-1)}$ and $\hat{\bm{G}}_x^{(t)}$, we concatenate $\bm{X}^{(t-1)}$ and $\hat{\bm{G}}_x^{(t)}$ as the input of the proximal network $\operatorname{proxNet}_{\theta_x^{(t)}}\left(\cdot\right)$, which introduces more flexibility to the combination of $\bm{X}^{(t-1)}$ and $\hat{\bm{G}}_x^{(t)}$.

% Compared with using gradient descent, a lot of iterative steps are usually required, we concatenate $\bm{X}^{(t-1)}$ and $\hat{\bm{G}}_x^{(t)}$ into the network and let the network learn a more flexible descent method for efficient convergence. 

\noindent
\textbf{Remark for K-net.} The proposed KXNet has clear physical interpretability. Different from the current most deep blind SR methods which heuristically adopt the concatenation or affine transformation operators on $\bm{K}^{(t-1)}$ to help the learning of HR images, the proposed K-net is constructed based on the iterative rule in  \cref{eq:important5} and every network sub-module has its specific physical meanings as shown in \cref{fig:fig2_b}. Specifically, following the degradation model, $\bm{X}^{(t-1)}$ is convolved with $\bm{K}^{(t-1)}$ followed by the downsampling operator. Then by subtracting the result from $\bm{Y}$, we get the residual information $\bm{e}_k^{(t)}$, which is actually the key information for updating the current estimation. Then, we regard $\bm{e}_k^{(t)}$ as a weight and adopt it to perform a weighted summation on the corresponded patches in $\bm{X}^{(t-1)}$ (\textit{i.e.}, 
row vectors in $D_\mathbf{s} U_f(\bm{X}^{(t-1)})\in\mathbb{R}^{hw\times p^{2}}$, as shown in \cref{fig:fig2_b}), and get $\bm{G}^{(t)}_k$ for updating $\bm{K}^{(t-1)}$. Actually, this is consistent with the relationship between $\bm{K}^{(t-1)}$ and $\bm{X}^{(t-1)}$, since convolution operation is executed based on the patch with $p\times p$.

\noindent
\textbf{Gradient adjuster.} For X-net, an adjuster is adopted to the gradient $\bm{G}_x^{(t)}$ as shown in the third equation of \cref{eq:important10}. Specifically, the transposed convolution $\bm{G}_x^{(t)}$ in X-net can easily cause ``uneven overlap'', putting more of the metaphorical paint in some places than others \cite{dumoulin2016guide,shi2016deconvolution}, which is unfriendly to the reconstruction of HR images. To alleviate the unevenness issue, we introduce the operator $\bm{K}^{(t)} \otimes_{\mathbf{s}}^{\mathrm{T}} \bm{1}$ and the adjusted gradient $\hat{\bm{G}}_x^{(t)}$ is\footnote{More analysis is provided in the supplementary material.}:
\begin{equation}\label{adjuster}
%   \vspace{-2mm}
  \hat{\bm{G}}_x^{(t)} = \frac{\bm{G}_x^{(t)}}{\bm{K}^{(t)} \otimes_{\mathbf{s}}^{\mathrm{T}} {\textcolor{black}{\bm{1}}}},
  \vspace{-2mm}
\end{equation}
where ${\bm{1}} \in \mathbb{R}^{h\times w}$ is a matrix with all elements as 1.

\vspace{-4mm}
\subsection{Network Training}
To train the proposed deep unfolding blind SR network, we utilize the $L_1$ loss \cite{zhao2016loss} to supervise the predicted blur kernel $\bm{K}^{(t)}$ and the estimated HR image $\bm{X}^{(t)}$ at each stage. Correspondingly, the total objective function is:
\vspace{-2mm}
\begin{equation}
    % \vspace{-2mm}
    L = \sum_{t=1}^{T} \alpha_{t} \big \| \bm{K} - \bm{K}^{(t)}\big \|_{1} + \sum_{t=1}^{T} \beta_{t} \big \| \bm{X} - \bm{X}^{(t)}\big\|_{1},
    \label{eq:important11}
    \vspace{-2mm}
\end{equation}
where $\bm{K}^{(t)}$ and $\bm{X}^{(t)}$ are obtained based on the updating process in \cref{eq:important9} and \cref{eq:important10} at the $t^{\text{th}}$ stage, respectively; $\alpha_{t}$ and $\beta_{t}$ are trade-off parameters \footnote{We set $\alpha_{t}=\beta_{t}=0.1$ at middle stages, $\alpha_{T}=\beta_{T}=1$ at the last stage, and $T=19$.}. $\bm{X}^{(0)}$ is initialized as the bicubic upsampling of the LR image $\bm{Y}$, and $\bm{K}^{(0)}$ is initialized as a standard Gaussian kernel.% More analysis is included in supplementary file.

\section{Experimental Results}\vspace{-2mm}
\subsection{Details Descriptions}%\vspace{-2mm}
% Based on \cref{eq:important}, we synthesize the image pairs for training and testing. 

\noindent\textbf{Synthesized Datasets.} Following \cite{gu2019blind,luo2020unfolding,wang2021unsupervised}, we collect 800 HR images from DIV2K \cite{agustsson2017ntire} and 2650 HR images from Flickr2K \cite{timofte2017ntire} to synthesize the training data, and adopt the four commonly-used benchmark datasets, \emph{i.e.}, Set5 \cite{bevilacqua2012low}, Set14 \cite{zeyde2010single}, BSD100 \cite{martin2001database}, and Urban100 \cite{huang2015single}, to synthesize the testing data. During the synthesis process of training and testing pairs, we adopt the degradation process in \cref{eq:important} with two different degradation settings: 1) isotropic Gaussian blur kernel with noise free; 2) anisotropic Gaussian blur kernel with noise~\cite{efrat2013accurate,riegler2015conditioned,zhang2020deep,liang2021flow}, and set the s-fold downsampler as in \cite{bell2019blind,zhang2020deep,liang2021flow}. Note that as stated in \cite{riegler2015conditioned,zhang2020deep}, the later setting is very close to the real SISR scenario.

In setting 1), for training set, following \cite{gu2019blind,luo2020unfolding,wang2021unsupervised}, the blur kernel size  $p\times p$ is set as $21 \times 21$ for all scales $\mathbf{s} \in \{2, 3, 4\}$ and the corresponding kernel width for different scales ($\times 2$, $\times 3$, and $\times 4$ SR) is uniformly sampled from the ranges $[0.2, 2.0]$, $[0.2, 3.0]$, and $[0.2, 4.0]$, respectively. For testing set, the blur kernel is set as \textit{Gaussian8} \cite{gu2019blind}, which uniformly samples 8 kernels from the ranges $[0.8, 1.6]$, $[1.35, 2.40]$, and $[1.8, 3.2]$ for the scale factors 2, 3, and 4, respectively.

In setting 2), for trainings set, we set the kernel size $p$ as $11/15/21$ for  $\times 2/3/4$ SR, respectively. Specifically,  the kernel width at each axis are obtained by randomly rotating the widths $\lambda_1$ and $\lambda_2$ with an angle $\theta \sim U[-\pi, \pi]$, where $\lambda_1$ and $\lambda_2$ are uniformly distributed in $U(0.6, 5.0)$. Besides, the range of noise level $\sigma$ is set to $[0, 25]$. For testing set, we separately set the kernel width as $\lambda_1=0.8, \lambda_2=1.6$ and $\lambda_1=2.0, \lambda_2=4.0$, and rotate them by $\theta\in\{0, \frac{\pi}{4}, \frac{\pi}{2}, \frac{3 \pi}{4}\}$, respectively. This means every HR testing image is degraded by 8 different blur kernels. 

\noindent\textbf{Real Dataset.} To verify the performance of the proposed method in real scenarios, we use the dataset RealSRSet \cite{zhang2021designing} for generalization evaluation, which includes 20 real LR images collected from various sources \cite{ignatov2017dslr,martin2001database,matsui2017sketch,zhang2018ffdnet}.

\noindent\textbf{Training Details.}  Based on the PyTorch framework executed on two RTX2080Ti GPUs, we adopt the Adam solver \cite{kingma2014adam} with the parameters as $\beta_1=0.9$ and $\beta_2=0.99$ to optimize the proposed network with the batch size and patch size set as 12 and $64\times 64$, respectively. The learning rate is initialized as $2\times 10^{-4}$ and decays by multiplying a factor of $0.5$ at every $2\times 10^5$ iteration. The training process ends when the learning rate decreases to $1.25 \times 10^{-5}$. 

%-------------------------------------------------------------------------
\begin{table*}[!ht]
    \vspace{-4mm}
    \scriptsize
    \renewcommand\arraystretch{1.0}
      \centering
            \caption{Average PSNR/SSIM of all the comparing methods \textbf{(Setting 1)}.} %\vspace{-2mm}
    \centering \setlength{\tabcolsep}{5.0pt}
      \begin{tabular}{l c c c c c c c c c c}
        \toprule
        % \hline
        \multirow{2}{*}{Method} & \multirow{2}{*}{Scale} & \multicolumn{2}{c}{Urban100 \cite{huang2015single}} & \multicolumn{2}{c}{BSD100 \cite{martin2001database}} & \multicolumn{2}{c}{Set14 \cite{zeyde2010single}} & \multicolumn{2}{c}{Set5 \cite{bevilacqua2012low}} \\
         & & PSNR & SSIM & PSNR & SSIM & PSNR & SSIM & PSNR & SSIM \\
        % \midrule
        \midrule
        % \hline
        % \hline
        Bicubic & \multirow{6}{*}{x2} & 24.23 & 0.7430 & 27.02 & 0.7472 & 27.13 & 0.7797 & 29.42 & 0.8666\\
        RCAN \cite{zhang2018image} & & 24.69 & 0.7727 & 27.40  & 0.7710 & 27.54 & 0.8804 & 29.81 & 0.8797 \\
        IKC \cite{gu2019blind} & & 29.22 & 0.8752 & 30.51 & 0.8540 & 31.69 & 0.8789 & 34.31 & 0.9287 \\
        DASR \cite{wang2021unsupervised} & & 30.63 & 0.9079 & 31.76 & 0.8901 & 32.93 & 0.9029 & 37.22 & 0.9515 \\
        DAN \cite{luo2020unfolding} & & 31.31 & 0.9165 & 31.93 & 0.8906 & 33.31 & 0.9085 & 37.54 & 0.9546 \\
        KXNet(ours) & & \bf 31.48 & \bf 0.9192 & \bf 32.03 & \bf 0.8941 & \bf 33.36 & \bf 0.9091 & \bf 37.58 & \bf 0.9552 \\
        % \midrule
        \hline
        Bicubic & \multirow{6}{*}{x3} & 22.07 & 0.6216 & 24.93 & 0.6360 & 24.58 & 0.6671 & 26.19 & 0.7716 \\
        RCAN \cite{zhang2018image} & & 22.18 & 0.6366 & 25.06 & 0.6501 & 24.73 & 0.6800 & 26.37 & 0.7840 \\
        IKC \cite{gu2019blind} & & 26.85 & 0.8087 & 28.29 & 0.7724 & 29.41 & 0.8106 & 32.90 & 0.8997 \\
        DASR \cite{wang2021unsupervised} & & 27.28 & 0.8307 & 28.85 & 0.7932 & 29.94 & 0.8266 & 33.78 & 0.9200 \\
        DAN \cite{luo2020unfolding} & & 27.94 & 0.8450 & 29.04 & 0.8001 & 30.24 & \bf 0.8350 & 34.18 & 0.9237 \\
        KXNet(ours) & & \bf 28.00 & \bf 0.8457 & \bf 29.06 & \bf 0.8010 & \bf 30.27 & 0.8340 & \bf 34.22 & \bf 0.9238 \\
        % \midrule
        \hline
        Bicubic & \multirow{6}{*}{x4} & 20.96 & 0.5544 & 23.84 & 0.5780 & 23.25 & 0.6036 & 24.43 & 0.7045 \\
        RCAN \cite{zhang2018image} & & 20.96 & 0.5608 & 23.89 & 0.5865 & 23.30 & 0.6109 & 24.52 & 0.7148 \\
        IKC \cite{gu2019blind} & & 24.42 & 0.7112 & 26.55 & 0.6867 & 26.88 & 0.7301 & 29.83 & 0.8375 \\
        DASR \cite{wang2021unsupervised} & & 25.49 & 0.7621 & 27.40 & 0.7238 & 28.26 & 0.7668 & 31.68 & 0.8854 \\
        DAN \cite{luo2020unfolding} & & 25.95 & 0.7787 & 27.53 & 0.7311 & 28.55 & 0.7749 & \bf 31.96 & 0.8898 \\
        KXNet(ours) & & \bf 26.18 & \bf 0.7873 & \bf 27.59 & \bf 0.7330 & \bf 28.67 & \bf 0.7782 & 31.94 & \bf 0.8912 \\
        \bottomrule
        % \hline
      \end{tabular}
    \vspace{-0.1cm}
      \label{tab:table1}
\end{table*}
%-------------------------------------------------------------------------

\begin{figure*}[!htp]
\centering
\subfigure{
    \begin{minipage}[b]{0.14\linewidth}
    \centerline{GT}
    \centerline{\includegraphics[width=2.0cm]{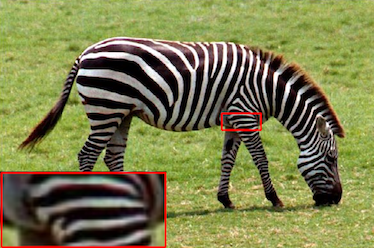}}
    \centerline{}
    \end{minipage}
    } %[图片大小]{图片路径}
\subfigure{
    \begin{minipage}[b]{0.14\linewidth}
    \centerline{Zoomed LR}
    \centerline{\includegraphics[width=2.0cm]{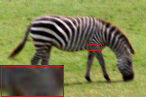}}
    \centerline{PSNR/SSIM}
    \end{minipage}
    } %[图片大小]{图片路径}
\subfigure{
    \begin{minipage}[b]{0.14\linewidth}
    \centerline{IKC \cite{gu2019blind}}
    \centerline{\includegraphics[width=2.0cm]{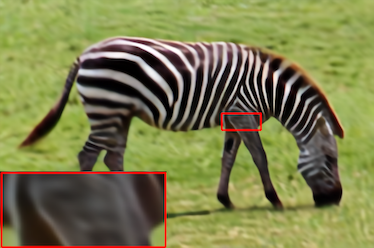}}
    \centerline{23.70/0.6281}
    \end{minipage}
    } %[图片大小]{图片路径}
\subfigure{
    \begin{minipage}[b]{0.14\linewidth}
    \centerline{DASR \cite{wang2021unsupervised}}
    \centerline{\includegraphics[width=2.0cm]{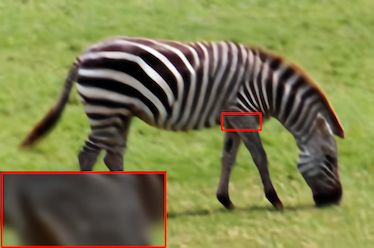}}
    \centerline{22.14/0.5794}
    \end{minipage}
    } %[图片大小]{图片路径}
\subfigure{
    \begin{minipage}[b]{0.14\linewidth}
    \centerline{DAN \cite{luo2020unfolding}}
    \centerline{\includegraphics[width=2.0cm]{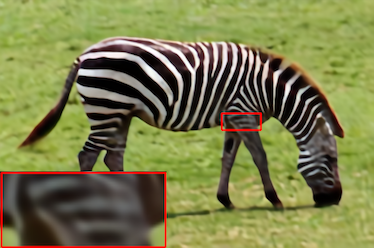}}
    \centerline{24.58/0.6567}
    \end{minipage}
    } %[图片大小]{图片路径}
\subfigure{
    \begin{minipage}[b]{0.14\linewidth}
    \centerline{{\bf KXNet}(ours)}
    \centerline{\includegraphics[width=2.0cm]{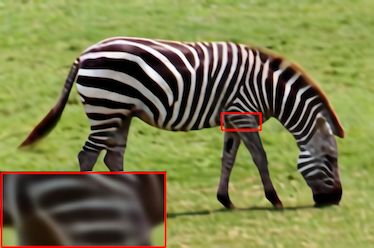}}
    \centerline{\textbf{24.87}/\textbf{0.6636}}
    \end{minipage}
    } %[图片大小]{图片路径}

\vspace{-3mm}
\subfigure{
    \begin{minipage}[b]{0.14\linewidth}
    % \centerline{GT}
    \centerline{\includegraphics[width=2.0cm]{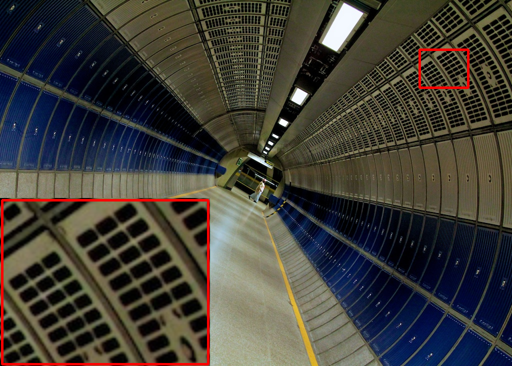}}
    \centerline{}
    \end{minipage}
    } %[图片大小]{图片路径}
\subfigure{
    \begin{minipage}[b]{0.14\linewidth}
    % \centerline{Zoomed LR}
    \centerline{\includegraphics[width=2.0cm]{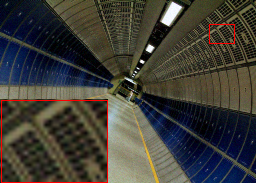}}
    \centerline{PSNR/SSIM}
    \end{minipage}
    } %[图片大小]{图片路径}
\subfigure{
    \begin{minipage}[b]{0.14\linewidth}
    % \centerline{IKC \cite{gu2019blind}}
    \centerline{\includegraphics[width=2.0cm]{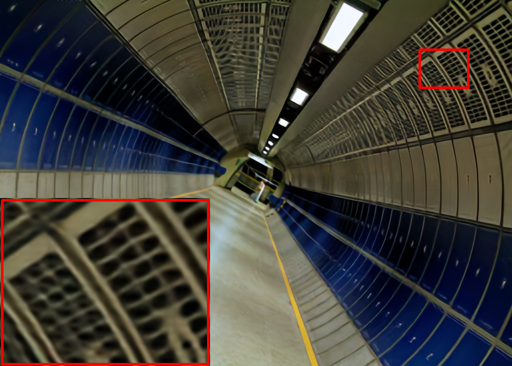}}
    \centerline{26.54/0.7391}
    \end{minipage}
    } %[图片大小]{图片路径}
\subfigure{
    \begin{minipage}[b]{0.14\linewidth}
    % \centerline{DASR \cite{wang2021unsupervised}}
    \centerline{\includegraphics[width=2.0cm]{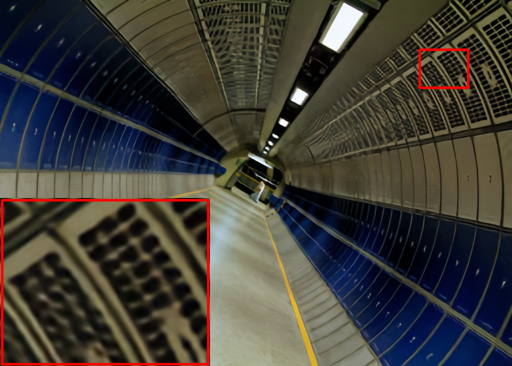}}
    \centerline{26.36/0.7280}
    \end{minipage}
    } %[图片大小]{图片路径}
\subfigure{
    \begin{minipage}[b]{0.14\linewidth}
    % \centerline{DAN \cite{luo2020unfolding}}
    \centerline{\includegraphics[width=2.0cm]{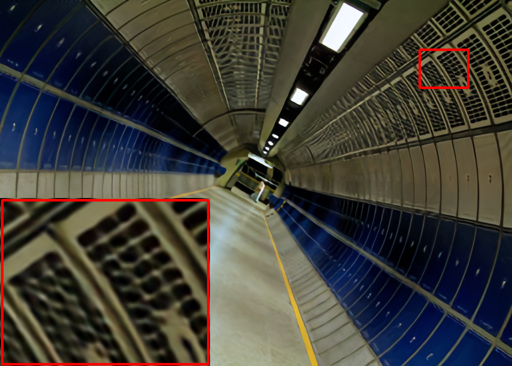}}
    \centerline{26.39/0.7389}
    \end{minipage}
    } %[图片大小]{图片路径}
\subfigure{
    \begin{minipage}[b]{0.14\linewidth}
    % \centerline{{\bf KXNet}(ours)}
    \centerline{\includegraphics[width=2.0cm]{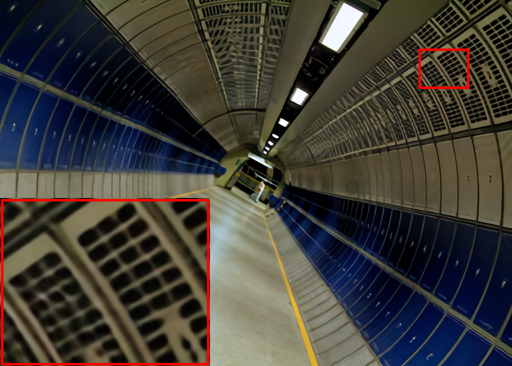}}
    \centerline{\textbf{26.85}/\textbf{0.7524}}
    \end{minipage}
    } %[图片大小]{图片路径}
% \subfigure[Jackson Yee]{\includegraphics[width=3.5cm]{image/014GT.png}}
% \subfigure[Jackson Yee]{\includegraphics[width=3.5cm]{image/014GT.png}}
\vspace{-2mm}
\caption{Performance comparison on \textit{img 14} in Set14 \cite{zeyde2010single} and \textit{img 078} in Urban100 \cite{huang2015single}. The scale factor is 4 and noise level is 5.} %图片标题
\label{fig:figure3}  %图片交叉引用时的标签
\vspace{-4mm}
\end{figure*}
%-------------------------------------------------------------------------

\noindent\textbf{Comparison Methods.} We comprehensively substantiate the superiority of our method by comparing it with several recent SOTA methods, including the non-blind SISR method RCAN \cite{zhang2018image}, and blind SISR methods, including IKC \cite{gu2019blind}, DASR \cite{wang2021unsupervised}, and DAN \cite{luo2020unfolding}.  For a fair comparison, we have retrained IKC, DASR, and DAN based on the aforementioned two settings with the public codes.

\begin{table}[!t]
\vspace{-6mm}
\scriptsize
% \small
% \renewcommand\arraystretch{0.90}
  \centering % \setlength{\tabcolsep}{40pt}
    \caption{Average PSNR/SSIM of all the comparing methods \textbf{(Setting 2)}.}
    \centering \setlength{\tabcolsep}{3.5pt}
\begin{tabular}{l c c c c c c c c c c}
    \toprule
    % \hline
     \multirow{2}{*}{Method} & \multirow{2}{*}{Scale} & \multicolumn{2}{c}{Urban100 \cite{huang2015single}} & \multicolumn{2}{c}{BSD100 \cite{martin2001database}} & \multicolumn{2}{c}{Set14 \cite{zeyde2010single}} & \multicolumn{2}{c}{Set5 \cite{bevilacqua2012low}}&Noise  \\
     & & PSNR & SSIM & PSNR & SSIM & PSNR & SSIM & PSNR & SSIM&Level \\
    % \midrule
    \midrule
    % \hline
    % \hline
     Bicubic & \multirow{6}{*}{x2} & 23.00 & 0.6656 & 25.85 & 0.6769 & 25.74 & 0.7085 & 27.68 & 0.8047 &\multirow{18}{*}{0} \\
     RCAN \cite{zhang2018image} & & 23.22 & 0.6791 & 26.03  & 0.6896 & 25.92 & 0.7217 & 27.85 & 0.8095 &\\
     IKC \cite{gu2019blind} & & 27.46 & 0.8401 & 29.85 & 0.8390 & 30.69 & 0.8614 & 33.99 & 0.9229 &\\
     DASR \cite{wang2021unsupervised} & & 26.65 & 0.8106 & 28.84 & 0.7965 & 29.44 & 0.8224 & 32.50 & 0.8961 & \\
     DAN \cite{luo2020unfolding} & & 27.93 & 0.8497 & 30.09 & 0.8410 & 31.03 & 0.8647 & 34.40 & 0.9291& \\
     KXNet(ours) & & \bf 28.33 & \bf 0.8627 & \bf 30.21 & \bf 0.8456 & \bf 31.14 & \bf 0.8672 & \bf 34.59 & \bf 0.9315 &\\
    % \midrule
    \cline{1-10}
     Bicubic & \multirow{6}{*}{x3} & 21.80 & 0.6084 & 24.68 & 0.6254 & 24.28 & 0.6546 & 25.78 & 0.7555& \\
     RCAN \cite{zhang2018image} & & 21.38 & 0.6042 & 24.47 & 0.6299 & 24.07 & 0.6606 & 25.63 & 0.7572 &\\
     IKC \cite{gu2019blind} & & 25.36 & 0.7626 & 27.56 & 0.7475 & 28.19 & 0.7805 & 31.60 & 0.8853& \\
     DASR \cite{wang2021unsupervised} & & 25.20 & 0.7575 & 27.39 & 0.7379 & 27.96 & 0.7727 & 30.91 & 0.8723& \\
     DAN \cite{luo2020unfolding} & & 25.82 & 0.7855 & 27.88 & 0.7603 & 28.69 & 0.7969 & 31.70 & 0.8940& \\
     KXNet(ours) & & \bf 26.37 & \bf 0.8035 & \bf 28.15 & \bf 0.7672 & \bf 29.04 & \bf 0.8036 & \bf 32.53 & \bf 0.9034& \\
    % \midrule
    \cline{1-10}
     Bicubic & \multirow{6}{*}{x4} & 20.88 & 0.5602 & 23.75 & 0.5827 & 23.17 & 0.6082 & 24.35 & 0.7086& \\
     RCAN \cite{zhang2018image} & & 19.84 & 0.5307 & 23.10 & 0.5729 & 22.38 & 0.5967 & 23.72 & 0.6973& \\
     IKC \cite{gu2019blind} & & 24.33 & 0.7241 & 26.49 & 0.6968 & 27.04 & 0.7398 & 29.60 & 0.8503& \\
     DASR \cite{wang2021unsupervised} & & 24.20 & 0.7150 & 26.43 & 0.6903 & 26.89 & 0.7306 & 29.53 & 0.8455& \\
     DAN \cite{luo2020unfolding} & & 24.91 & 0.7491 & 26.92 & 0.7168 & 27.69 & 0.7600 & 30.53 & 0.8746& \\
     KXNet(ours) & & \bf 25.30 & \bf 0.7647 & \bf 27.08 & \bf 0.7221 & \bf 27.98 & \bf 0.7659 & \bf 30.99 & \bf 0.8815& \\
    % \bottomrule
    \hline

     Bicubic & \multirow{6}{*}{x2} & 22.19 & 0.5159 & 24.44 & 0.5150 & 24.38 & 0.5497 & 25.72 & 0.6241& \multirow{18}{*}{15} \\
     RCAN \cite{zhang2018image} & & 21.28 & 0.3884 & 22.98  & 0.3822 & 22.96 & 0.4155 & 23.76 & 0.4706& \\
     IKC \cite{gu2019blind} & & 24.69 & 0.7208 & 26.49 & 0.6828 & 26.93 & 0.7244 & 29.21 & 0.8260& \\
     DASR \cite{wang2021unsupervised} & & 24.84 & 0.7273 & 26.63 & 0.6841 & 27.22 & 0.7283 & 29.44 & 0.8322&  \\
     DAN \cite{luo2020unfolding} & & 25.32 & 0.7447 & 26.84 & 0.6932 & 27.56 & 0.7392 & 29.91 & 0.8430& \\
     KXNet(ours) & & \bf 25.45 & \bf 0.7500 & \bf 26.87 & \bf 0.6959 & \bf 27.59 & \bf 0.7422 & \bf 29.93 & \bf 0.8449& \\
    % \midrule
    % \midrule
    \cline{1-10}
     Bicubic & \multirow{6}{*}{x3} & 21.18 & 0.4891 & 23.55 & 0.4961 & 23.28 & 0.5289 & 24.42 & 0.6119& \\
     RCAN \cite{zhang2018image} & & 20.22 & 0.3693 & 22.20 & 0.3726 & 21.99 & 0.4053 & 22.85 & 0.4745& \\
     IKC \cite{gu2019blind} & & 24.21 & 0.7019 & 25.93 & 0.6564 & 26.42 & 0.7018 & 28.61 & 0.8135& \\
     DASR \cite{wang2021unsupervised} & & 23.93 & 0.6890 & 25.82 & 0.6484 & 26.27 & 0.6940 & 28.27 & 0.8047& \\
     DAN \cite{luo2020unfolding} & & 24.17 & 0.7013 & 25.93 & 0.6551 & 26.46 & 0.7014 & 28.52 & 0.8130& \\
     KXNet(ours) & & \bf 24.42 & \bf 0.7135 & \bf 25.99 & \bf 0.6585 & \bf 26.56 & \bf 0.7063 & \bf 28.64 & \bf 0.8178& \\
    % \midrule
    \cline{1-10}
     Bicubic & \multirow{6}{*}{x4} & 20.38 & 0.4690 & 22.83 & 0.4841 & 22.39 & 0.5120 & 23.33 & 0.5977& \\
     RCAN \cite{zhang2018image} & & 19.23 & 0.3515 & 21.47 & 0.3686 & 21.05 & 0.3960 & 21.77 & 0.4689& \\
     IKC \cite{gu2019blind} & & 23.35 & 0.6665 & 25.21 & 0.6238 & 25.58 & 0.6712 & 27.45 & 0.7867& \\
     DASR \cite{wang2021unsupervised} & & 23.26 & 0.6620 & 25.20 & 0.6223 & 25.55 & 0.6683 & 27.32 & 0.7842& \\
     DAN \cite{luo2020unfolding} & & 23.48 & 0.6742 & 25.25 & 0.6283 & 25.72 & 0.6760 & 27.55 & 0.7938& \\
     KXNet(ours) & & \bf 23.67 & \bf 0.6844 & \bf 25.30 & \bf 0.6296 & \bf 25.78 & \bf 0.6792 & \bf 27.66 & \bf 0.7977&\\

    \bottomrule
    % \hline
  \end{tabular}
%   \vspace{4mm}
  \label{tab:table2}
  \vspace{-4mm}
\end{table}

\vspace{-1mm}
\noindent\textbf{Performance Evaluation.} For synthetic data, we adopt the PSNR and SSIM \cite{wang2004image} computed on  Y channel in  YCbCr space. While for RealSRSet, we only provide the visual results since there is no ground-truth (GT) image.

\subsection{Experiments on Synthetic Data}
%\vspace{-1mm}
\cref{tab:table1} reports the average PSNR and SSIM of all the comparison methods on four benchmark datasets with the first simple synthesis setting. From it, we can easily find that the proposed KXNet is superior or at least comparable to other comparison methods under different scales. This is mainly attributed to its proper and full embedding of physical generation mechanism which finely helps the KXNet to be trained in the right direction. 

% In addition, although RCAN achieved significant performance in the case of bicubic downsampling, it suffers from severe performance drop when kernels deviate from the assumption. IKC trains multiple models separately, which will cause small errors in the front module to be magnified, resulting performance drop. By introducing contrastive learning to learn degradation representations, DASR is still limited to feature learning. DAN as a completely end-to-end method dose achieve remarkable performance. However, the improper use of kernel information restricts its performance further. Our KXNet fully considers the physical generation mechanism, which can guide the end-to-end unfolding network to achieve better results. 

\cref{tab:table2} provides the quantitative comparison where testing sets are synthesized under the second complicated setting. Clearly, even in this hard case, our proposed KXNet still achieves the most competing performance and consistently outperforms other comparison methods on the four benchmark datasets with different SR scales. This comprehensively substantiates the generality  of the proposed method and it potential usefulness in real-world scenarios. 

\cref{fig:figure3} visually displays the SR results on \textit{img014} from Set14 and \textit{img078} from Urban100 where the corresponding LR images are synthesized based on the second settings. As seen, almost all the blind comparison methods cannot finely reconstruct the image details. However, our KXNet achieves superior SR performance and the SR images contain more useful textures and sharper edges.

\vspace{-4mm}
\subsection{More Analysis and Verification}
\noindent{\bf Number of Iterations for KXNet.} To explore the effectiveness of KXNet, we investigate the effect of the number of iterations on the performance of KXNet. In \cref{tab:stage_number}, $S=0$ means that the initialization $\bm{X}^{(0)}$ and $\bm{K}^{(0)}$ are directly used as the recovery result. Taking $S = 0$ as a baseline, we can clearly see that when $S=5$, our method has been able to achieve a significant recovery performance which strongly validates the effectiveness of K-net and X-net. Beside, since larger stages would make gradient propagation more difficult, the case $S=21$ has the same PSNR to the case $S=19$. Thus, we set $S=19$ for this work.

%-----------------

\begin{figure}[t]
\vspace{-3mm}
\centering
\includegraphics[width=\linewidth]{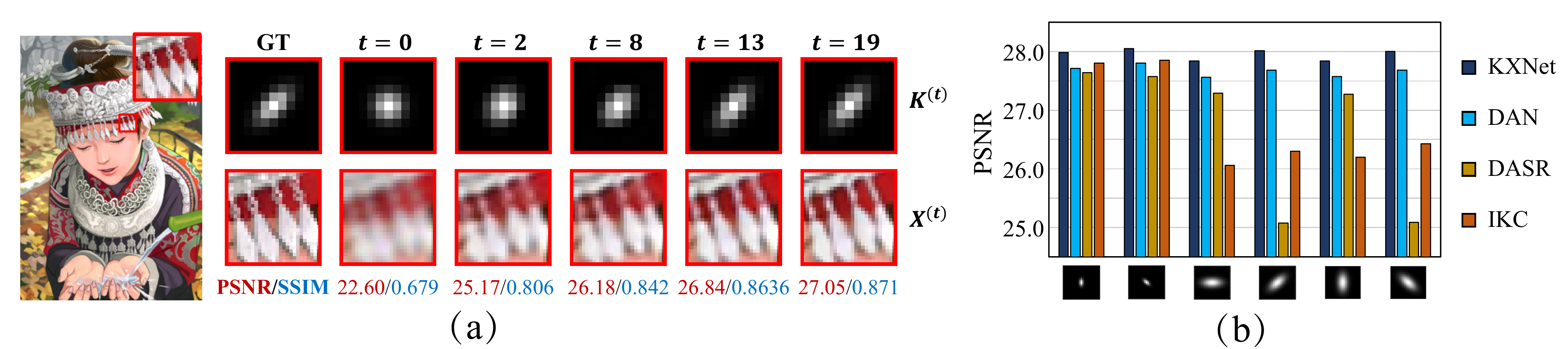}
 \vspace{-4mm}
\caption{(a) The estimated SR image and the extracted blur kernel at different iterative stages of KXNet. (b) Performance comparison under different blur kernel settings on Set14 \cite{zeyde2010single} (scale = 4, noise = 0).}
\label{fig:girl_ker}
 \vspace{-4mm}
\end{figure}

\begin{table}[t]
% \scriptsize
% \resizebox{\textwidth}{12mm}{}
\centering
  \caption{Effect of stage number S on the performance of KXNet on Set14.}
%   \vspace{-1mm}
  \centering\setlength{\tabcolsep}{9.0pt}
    % \centering
  \begin{tabular}{l  c  c  c  c  c  c}
    % \toprule
\toprule
    Stage No. & S=0 & S=5 & S=10 & S=17 & S=19 & S=21 \\
    \midrule
    PSNR & 25.74 & 29.91 & 30.57 & 30.96 & 31.14 & 31.14 \\

    SSIM & 0.7085 & 0.8400 & 0.8556 & 0.8631 & 0.8672 & 0.8665 \\

    Params(M) & - & 1.72 & 3.42 & 5.82 & 6.50 & 7.18 \\

    Speed(seconds) & - & 0.51 & 0.54 & 0.58 & 0.59 & 0.64 \\
    %\hline
    \bottomrule
  \end{tabular}
%   \caption{Effect of stage number S on the performance of KXNet.}
    \label{tab:stage_number}
\hfill
\vspace{-4mm}
\end{table}
%---------

\begin{table}[t]
\flushright\vspace{-3mm}
  \caption{Average inference speed of different methods on Set5.}
  \centering\setlength{\tabcolsep}{25.5pt}
  \begin{tabular}{l c c c}
    \toprule
    % \hline
    Methods & IKC \cite{gu2019blind} & DAN \cite{luo2020unfolding} & KXNet \\ 
    % \hline
    \midrule
    Times (s) & 2.15 & 0.52 & 0.38  \\
    % \hline
    % \hline
    \bottomrule
    % \bottomrule
  \end{tabular}
%   \caption{Effect of stage number S on the performance of KXNet.}
  \label{tab:table4}
\vspace{-1mm}
\end{table}
%-------------------------------------------------------

% \vspace{-2mm}
\noindent{\bf Non-Blind Super-Resolution.} We provide ground truth (GT) kernel to verify the effectiveness of KXNet and other method on Set14 \cite{zeyde2010single}. Providing blur kernel for KXNet and DAN, namely KXNet(GT kernel) and DAN(GT kernel), the PSNR of the results are 32.85 and 32.67, respectively. While the baseline KXNet with unknown blur kernel is 31.97. This means that when we provide X-net with an accurate blur kernel, the restoration accuracy can be further improved while illustrating the rationality and superiority of X-net for image restoration.

\noindent{\textbf{Stage Visualization.}}
Owning to the full embedding of the physical generation mechanism, the proposed KXNet can facilitate us to easily understand the iterative process via stage visualization. As shown in \cref{fig:girl_ker}(a), it presents the estimated SR images and the predicted blur kernel at different stages, where the blur kernel is simply initialized as standard Gaussian kernel. Clearly, with the increasing of the iterative stages, the extracted blur kernel has a better and clearer pattern, which is getting closer to the GT kernel. Correspondingly, the SR image is gradually ameliorated and achieves higher PSNR/SSIM scores. This interpretable learning process is the inherent characteristic of the proposed KXNet which is finely guided by the mutual promotion between K-net and X-net.

\vspace{1mm}
\noindent{\bf Robustness to Blur Kernel.} 
To comprehensively validate the effectiveness of the proposed method and its advantage over blur kernel extraction, we compare the SR performance of different methods on synthesized Set14~\cite{zeyde2010single} with different blur kernel widths. As displayed in \cref{fig:girl_ker}(b),  as the structures of the testing blur kernels become more complex, the performance of most comparison methods has severely deteriorated. However, the proposed KXNet can consistently achieve the most competing PSNR scores and the fluctuation is very small. This result fully shows that our method has better robustness to the types of blur kernels and it has better potential to deal with general and real scenes.

\vspace{-2mm}
\subsection{Inference Speed}
Based on Set5 under setting2 (scale=2, noise=0), we evaluate the inference time computed on an P100 GPU. For every image, the average testing time for IKC \cite{gu2019blind}, DAN \cite{luo2020unfolding}, and our proposed KXNet are shown in \cref{tab:table4}. Clearly, compared to these representative SOTA methods, our method has high inference speed and better computation efficiency, which is meaningful for real applications.
% One more superiority of our end-to-end model is that it has a higher inference speed. For quantitative comparison, we evaluated the average speed of different methods on the same platform (an P100 GPU), testset is the same as setting2 with 0 noise level and x2 SR in Set5. The average speed of IKC \cite{gu2019blind} is 2.15 seconds per image, and DAN \cite{luo2020unfolding} is  per image, and our method is  per image.

%------------------------------------------------------------

\begin{figure*}[t]
\centering
\subfigure{
    \begin{minipage}[b]{0.17\linewidth}
    \centerline{\includegraphics[width=2.3cm]{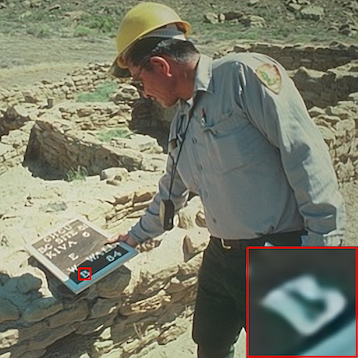}}
    \centerline{\includegraphics[width=2.3cm]{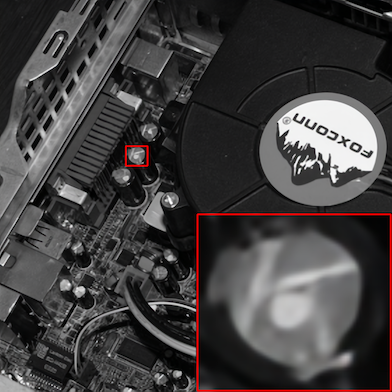}}
    \centerline{IKC \cite{gu2019blind}}
    \end{minipage}
    } %[图片大小]{图片路径}
\subfigure{
    \begin{minipage}[b]{0.17\linewidth}
    \centerline{\includegraphics[width=2.3cm]{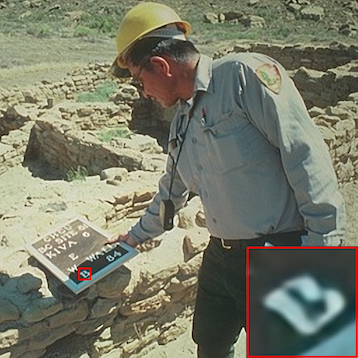}}
    \centerline{\includegraphics[width=2.3cm]{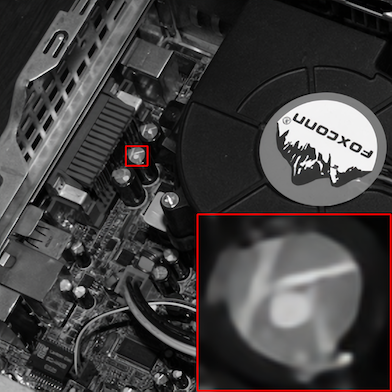}}
    \centerline{DASR \cite{wang2021unsupervised}}
    \end{minipage}
    } %[图片大小]{图片路径}
\subfigure{
    \begin{minipage}[b]{0.17\linewidth}
    \centerline{\includegraphics[width=2.3cm]{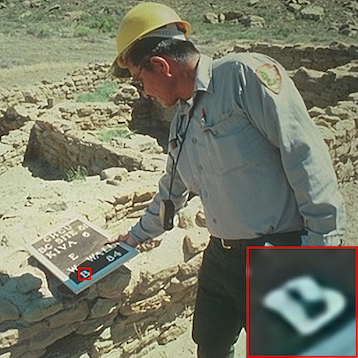}}
    \centerline{\includegraphics[width=2.3cm]{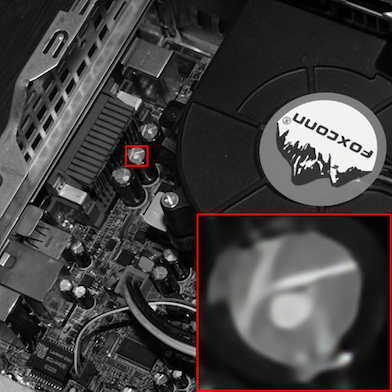}}
    \centerline{DAN \cite{luo2020unfolding}}
    \end{minipage}
    } %[图片大小]{图片路径}
\subfigure{
    \begin{minipage}[b]{0.17\linewidth}
    \centerline{\includegraphics[width=2.3cm]{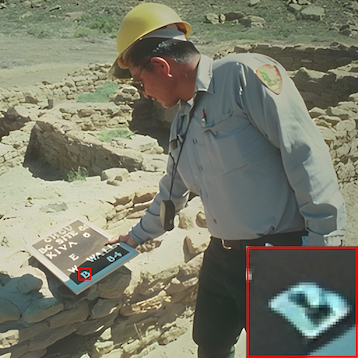}}
    \centerline{\includegraphics[width=2.3cm]{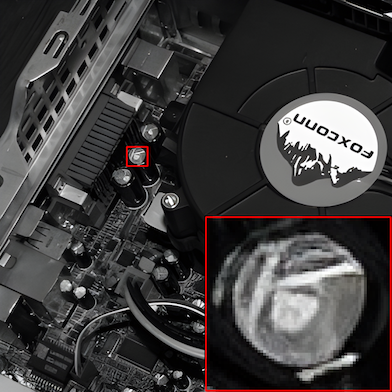}}
    \centerline{Real-ESRGAN \cite{wang2021real}}
    \end{minipage}
    } %[图片大小]{图片路径}
\subfigure{
    \begin{minipage}[b]{0.17\linewidth}
    \centerline{\includegraphics[width=2.3cm]{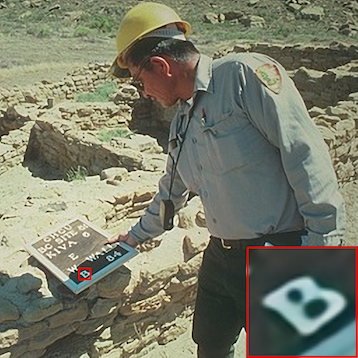}}
    \centerline{\includegraphics[width=2.3cm]{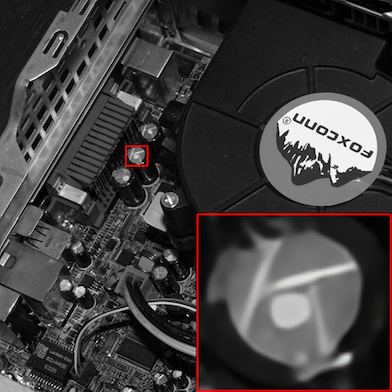}}
    \centerline{KXNet}
    \end{minipage}
    } %[图片大小]{图片路径}
% \subfigure[Jackson Yee]{\includegraphics[width=3.5cm]{image/014GT.png}}
% \subfigure[Jackson Yee]{\includegraphics[width=3.5cm]{image/014GT.png}}
\vspace{-2mm}
\caption{Visual comparison on RealSRSet with scale factor as 4.} %图片标题
\label{fig:realSR}  %图片交叉引用时的标签
\vspace{-6mm}
\end{figure*}
%--------------------------------------------------------------------------
%-------------------------------------------------------------------------

\vspace{-4mm}
\subsection{Experiments on Real Images}
We further evaluate the effectiveness of our method for real-world image restoration on RealSRSet \cite{zhang2021designing}. As shown in \cref{fig:realSR}, the proposed KXNet can recover clearer edges and generate more useful information.

\vspace{-2mm}
\section{Conclusion}
In this paper, we have proposed an end-to-end blind super-resolution network for SISR, named as KXNet. In specific, we analyze the classical degradation process of low-resolution (LR) images and utilize the proximal gradient technique to derive an optimization algorithm. By unfolding the iterative steps into network modules, we easily construct the entire framework which is composed of K-net and X-net, and the explicit physical generation mechanism of blur kernels and high-resolution (HR) images 
are fully incorporated into the entire learning process.  Besides, the proposed KXNet has better potential to finely extract different types of blur kernels which should be useful for other related tasks. All these advantages have been fully substantiated by comprehensive experiments executed on synthesized and real-world data under different degradation settings. This also finely validates the effectiveness and generality of the proposed KXNet. 

\vspace{2mm}
{\noindent\textbf{Acknowledgment}. This research was supported by NSFC project under contracts U21A6005, 61721002, U1811461, 62076196, The Major Key Project of PCL under contract PCL2021A12, and the Macao Science and Technology Development Fund under Grant 061/2020/A2.

\clearpage
\bibliographystyle{splncs04}
\bibliography{egbib}
\end{document}